%% file: main.tex
\pdfoutput=1

\documentclass[10pt,twocolumn,letterpaper]{article}

\newif\ifarxivversion
\arxivversiontrue

\ifarxivversion
  \usepackage[pagenumbers]{wacv}
\else
  \usepackage[review,datasets]{wacv}
\fi


\definecolor{wacvblue}{rgb}{0.21,0.49,0.74}
\ifarxivversion
  \usepackage[breaklinks,colorlinks,allcolors=wacvblue]{hyperref}
\else
  \usepackage[pagebackref,breaklinks,colorlinks,allcolors=wacvblue]{hyperref}
\fi

\def\confName{WACV}
\def\confYear{2027}
\ifarxivversion\else
  \def\wacvPaperID{*****}
\fi

\title{Interpretable Temporal Video Reasoning with EventGraph and EventField}

\ifarxivversion
  \author{Durgendra Narayan Singh\\
  \textit{Independent Researcher}\\
  Twin Falls, ID, USA\\
  {\tt\small durgendra@gmail.com}}
\else
  \author{Anonymous WACV submission}
\fi

\begin{document}
\maketitle

\input{sec/0_abstract}
\input{sec/1_intro}
\input{sec/2_related_work}
\input{sec/2_method}
\input{sec/3_experiments}
\input{sec/4_results}
\input{sec/5_conclusion}
\input{sec/6_limitations}

{
    \small
    \bibliographystyle{ieeenat_fullname}
    \bibliography{main}
}

\ifarxivversion
  \clearpage
  \appendix
  \input{sec/7_reproducibility_summary}
  \input{sec/6_appendix_eventglyphs}
\fi

\end{document}

%% file: sec/0_abstract.tex
\begin{abstract}
We present a structured temporal video reasoning pipeline built around a discrete EventGraph, a continuous EventField, and a human-readable EventGlyph view. On a calibrated EPIC-KITCHENS subset of 10 videos and 50 temporal reasoning questions, EventField+Glyph achieves 0.98 overall accuracy, which is higher than the caption baseline by +0.40 (paired $p = 1.1 \times 10^{-5}$) and direct VLM-only QA by +0.20 ($p = 0.0063$) on this subset. We further evaluate annotation-source variations, including manual, heuristic, and heuristic+Gemini pipelines, and find that the best structured method stays above the caption baseline across settings. We also include cross-video pair benchmarking and an appendix gallery of glyph outputs for all studied videos. Overall, the results indicate that structured temporal representations can support both performance and inspectability by preserving symbolic structure, capturing temporal continuity, and providing human-readable diagnostics for video reasoning.
\end{abstract}

%% file: sec/1_intro.tex
\section{Introduction}
\label{sec:intro}

Temporal reasoning from video is central to robotics, egocentric assistance, and broader video inference, reasoning, and understanding tasks. Captions are convenient, but they often collapse ordering, duration, overlap, and causality into a single textual summary, which makes them weak for questions that depend on temporal structure.

This paper studies a structured alternative for temporal video understanding. Prior work on compositional temporal reasoning, modular video QA, and grounded VideoQA has shown the value of explicit intermediate structure, temporal grounding, and inspectable evidence, while interpretability research argues for representations that can be inspected directly rather than inferred only from latent vectors \cite{lei2018tvqa,jang2017tgifqa,grunde-mclaughlin2021agqa,yu2023anetqa,min2024morevqa,xiao2024grounded,di2024groundvqa,li2024mvbench,wang2018videos,lipton2018mythos,rudin2019stop}. Our goal is not to replace those methods, but to combine symbolic events, continuous timing, and human-readable diagnostics into a practical pipeline for reasoning and review.

The proposed pipeline represents each clip with three complementary views. EventGraph captures symbolic event structure, EventField turns those events into a continuous time signal, and EventGlyph renders the resulting dynamics in a compact human-readable form. We evaluate this stack on a fixed EPIC-KITCHENS subset of 10 videos and 50 questions, with source-variation and cross-video pair tests designed to check robustness rather than benchmark scale. A formal human-subject study is left for future work, so the present paper treats interpretability claims as qualitative and evidence-based rather than user-validated.
In practice, this matters because the same clip can be checked from three angles: as a sequence of discrete events, as a temporal signal, and as a compact operator-facing summary. That makes it easier to identify whether a failure comes from event extraction, timing, or downstream reasoning, and it gives reviewers a clearer view of what the model is actually using.

The main contributions are:
\begin{enumerate}[leftmargin=1.2em]
\item An end-to-end, reproducible EventGraph, EventField, and EventGlyph pipeline.
\item A comparative evaluation against caption-only, VLM-only, event-graph, EventField, and EventField+Glyph variants.
\item A source-variation study spanning manual, heuristic, and heuristic+Gemini annotations, plus cross-video pair analysis.
\end{enumerate}

On this benchmark, the structured EventField+Glyph variant performs better than caption-only and is close to the event-graph baseline, while also providing explicit temporal diagnostics for human review. Because the evaluation set is intentionally small, we present the results as evidence for the representation's usefulness, not as a claim of broad benchmark supremacy.

%% file: sec/2_related_work.tex
\section{Related Work}
\label{sec:related_work}

\textbf{Temporal video reasoning and QA.}
Temporal QA benchmarks such as TVQA, TGIF-QA, AGQA, and ANetQA established that video reasoning is not just about recognizing content, but also about handling order, causality, compositional structure, and temporal localization \cite{lei2018tvqa,jang2017tgifqa,grunde-mclaughlin2021agqa,yu2023anetqa}. More recent benchmarks and models such as MVBench, MoReVQA, grounded VideoQA, and long-egocentric QA further show that explicit intermediate structure and temporal grounding remain difficult but important for modern video understanding systems \cite{li2024mvbench,min2024morevqa,xiao2024grounded,di2024groundvqa}. EPIC-KITCHENS provides a strong egocentric setting for dense action and temporal reasoning analysis \cite{damen2020epic}.

\textbf{Graph-based video structure.}
Graph-structured representations are widely used to model objects, relations, and action dependencies in video \cite{wang2018videos}. They help expose symbolic structure, but discrete graphs alone can miss continuous timing variation.

\textbf{Interpretability and process views.}
Interpretability-focused work argues for transparent intermediate structure rather than only dense black-box embeddings \cite{lipton2018mythos,rudin2019stop}. In process and workflow settings, visual traces are useful for diagnostics and compliance analysis \cite{vanderaalst2016process}.

\textbf{Positioning.}
Our EventGraph, EventField, and EventGlyph pipeline combines these ideas into a single evaluable stack for temporal reasoning and inspection. Unlike prior work that primarily optimizes answer accuracy or grounding alone, our focus is on a compact representation that preserves symbolic structure, continuous timing, and human-readable diagnostics in the same pipeline.

%% file: sec/2_method.tex
\section{Method}
\label{sec:method}

\subsection{EventGraph}
Each clip is represented by event annotations with attributes $(id, label, type, t_{start}, t_{end}, confidence)$ and typed temporal or causal edges. The result is a directed graph that preserves node and edge metadata for downstream reasoning.

\subsection{EventField}
For channel $c$ over normalized time $t \in [0,1]$, we define
\[
F_c(t) = \sum_i a_i \cdot K(t; \mu_i, \sigma_i),
\]
where $a_i$ is confidence-weighted amplitude, $\mu_i$ is the event midpoint, and $\sigma_i$ is duration-derived spread. We use Gaussian kernels by default, with triangular kernels as an alternative. Channels include event types plus optional causal-influence and uncertainty channels.

\subsection{EventGlyph}
EventField is projected onto a circular glyph: angle encodes time and radius encodes magnitude. Channel traces overlay the total activity contour, causal links appear as internal connectors, and uncertainty is represented as an outer halo. The glyph is intended for fast operator inspection and failure localization.
Taken together, the three views serve different purposes rather than duplicating one another. EventGraph provides discrete structure for reasoning over actions and relations, EventField preserves temporal continuity, and the glyph compresses the combined state into a form that is fast to inspect by eye. This division is what lets the pipeline remain interpretable while still supporting temporal QA.

\subsection{Compared variants}
We compare five method variants under the same task set:
\begin{enumerate}[leftmargin=1.2em]
\item caption-only baseline,
\item VLM-only baseline using video frames directly,
\item event-graph-only baseline,
\item EventField reasoning,
\item EventField + Glyph (Structured task-aware variant).
\end{enumerate}

\subsection{Annotation-source variations}
In addition to method ablations, we evaluate three annotation pipelines on the same 10 videos:
\begin{enumerate}[leftmargin=1.2em]
\item manual EPIC-derived annotations converted to the repository schema,
\item heuristic motion and state-change proposals from raw video,
\item heuristic proposals refined with VLM-assisted relabeling and confidence.
\end{enumerate}
For the heuristic-family runs, we use the repository-default EPIC10 calibration settings, including a 0.6 s merge gap, a 0.50 weak-state confidence threshold, a 1.2 s semantic window, and a maximum of 22 state changes per minute.
This isolates whether the reasoning stack remains robust as upstream annotation quality changes.

%% file: sec/3_experiments.tex
\section{Experiments}
\label{sec:experiments}

\subsection{Data and tasks}
We evaluate on a fixed calibrated EPIC-KITCHENS subset of 10 videos with 50 total QA items. The same question set and answer keys are used for every method, and the cross-video pair benchmark is a separate fixed task-v2 set with 36 matched QA items. We do not resample the evaluation split across runs.

Task types are order, before/after, causal, duration comparison, and next-event prediction.

\subsection{Evaluation protocol}
Overall accuracy is the fraction of exactly correct answers. Macro average is the unweighted mean of the per-task-type accuracies over order, before/after, causal, duration comparison, and next-event questions.

For pairwise comparisons, we use paired bootstrap confidence intervals with 2,000 resamples and seed 42, together with exact two-sided McNemar tests on matched predictions. Statistical conclusions are based on paired items only.

For annotation-source variants, we report the best method per source, absolute delta versus the caption baseline, significance of best-versus-caption, and annotation complexity statistics.
These choices are meant to keep the evaluation compact but still sensitive to both overall correctness and performance on specific temporal reasoning modes. The paired tests are appropriate because every method is evaluated on the same item set, which makes each answer a matched comparison rather than an independent sample. Reporting both overall and macro average also prevents the strongest task types from hiding weaknesses on rarer question categories.
The pruning settings used for auto-annotation were selected by a small sweep over predefined options, and we treat that choice as a tuning step rather than a separate benchmark result.

\subsection{Implementation details}
The VLM-only baseline samples video at 1.0 fps with a cap of 24 frames, and all method comparisons are scored against the same exported task manifests.

\subsection{Reproducibility}
All reported numbers were generated from version-controlled artifacts and exported into the tables and figures shown in this paper.

%% file: sec/4_results.tex
\section{Results}
\label{sec:results}

\subsection{Main comparison}
\input{tables/main_results}
Structured methods outperform the caption-only baseline in this subset. The strongest variant, EventField+Glyph, reaches 0.98 overall accuracy on the 50-item EPIC10 task set, while VLM-only reaches 0.78.
This gap is consistent with the idea that explicit temporal structure is most useful when the answer depends on ordering, duration, or state change rather than on coarse scene recognition. It also suggests that the glyph view may contribute diagnostic value within the structured stack, because the strongest result appears only once the structured reasoning variant is used.

\begin{figure}[t]
  \centering
  \includegraphics[width=0.95\linewidth]{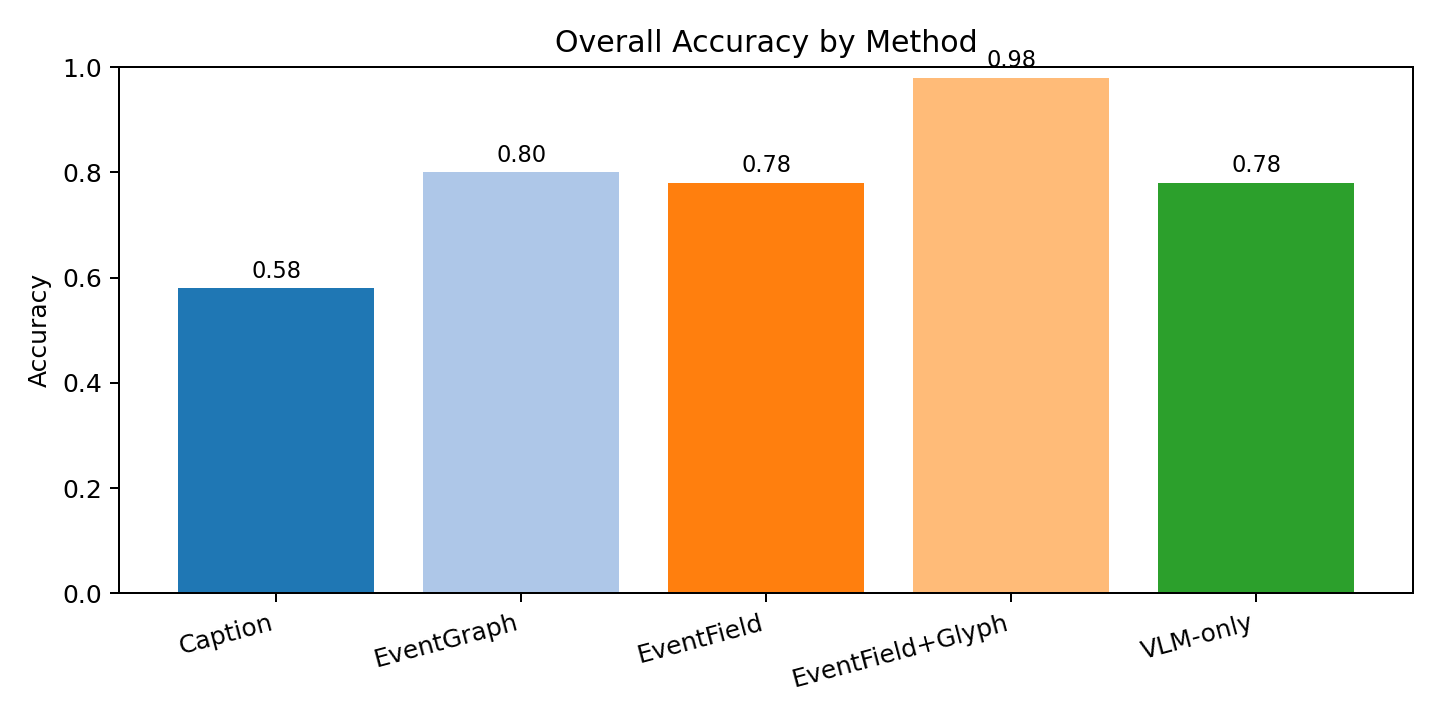}
  \caption{Overall accuracy comparison across methods.}
  \label{fig:overall}
\end{figure}

\subsection{Pairwise significance}
\input{tables/pairwise_significance}
The gains over caption are supported by paired tests on this subset. In particular, caption versus EventField+Glyph shows a large absolute gap, and the structured variant also performs better than direct VLM-only QA.

\begin{figure*}[t]
  \centering
  \includegraphics[width=0.98\textwidth]{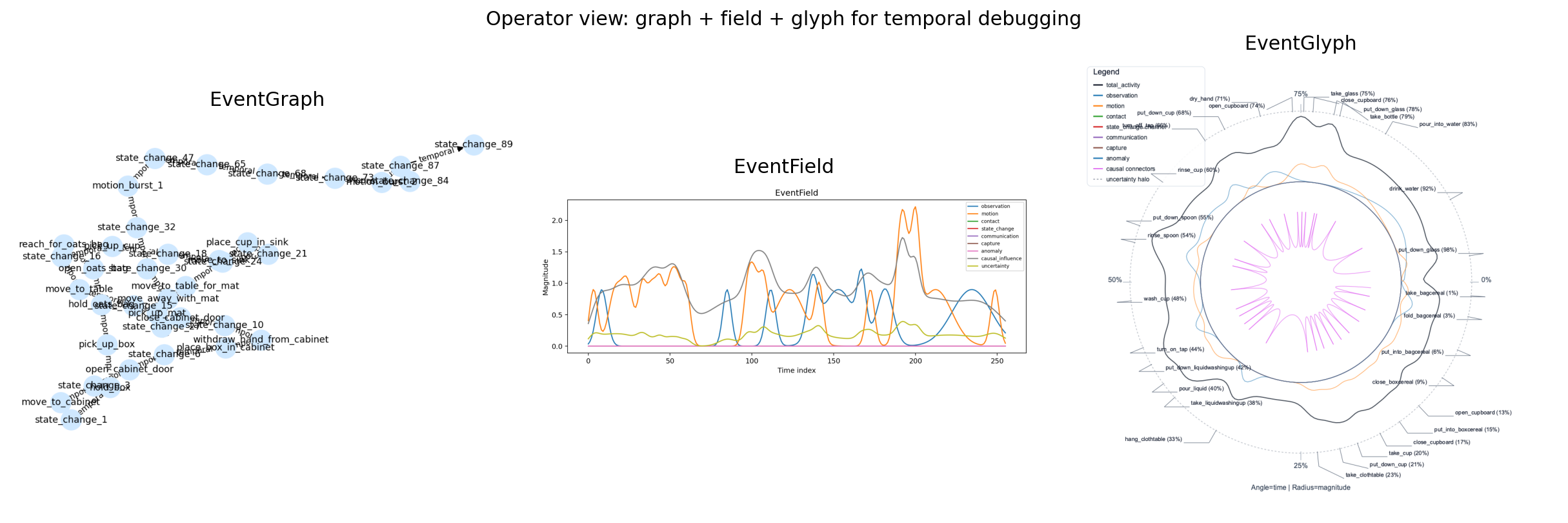}
  \caption{Operator view for debugging: event graph, EventField, and Glyph from one clip.}
  \label{fig:operator}
\end{figure*}

\subsection{Task-wise behavior}
\begin{figure}[t]
  \centering
  \includegraphics[width=0.95\linewidth]{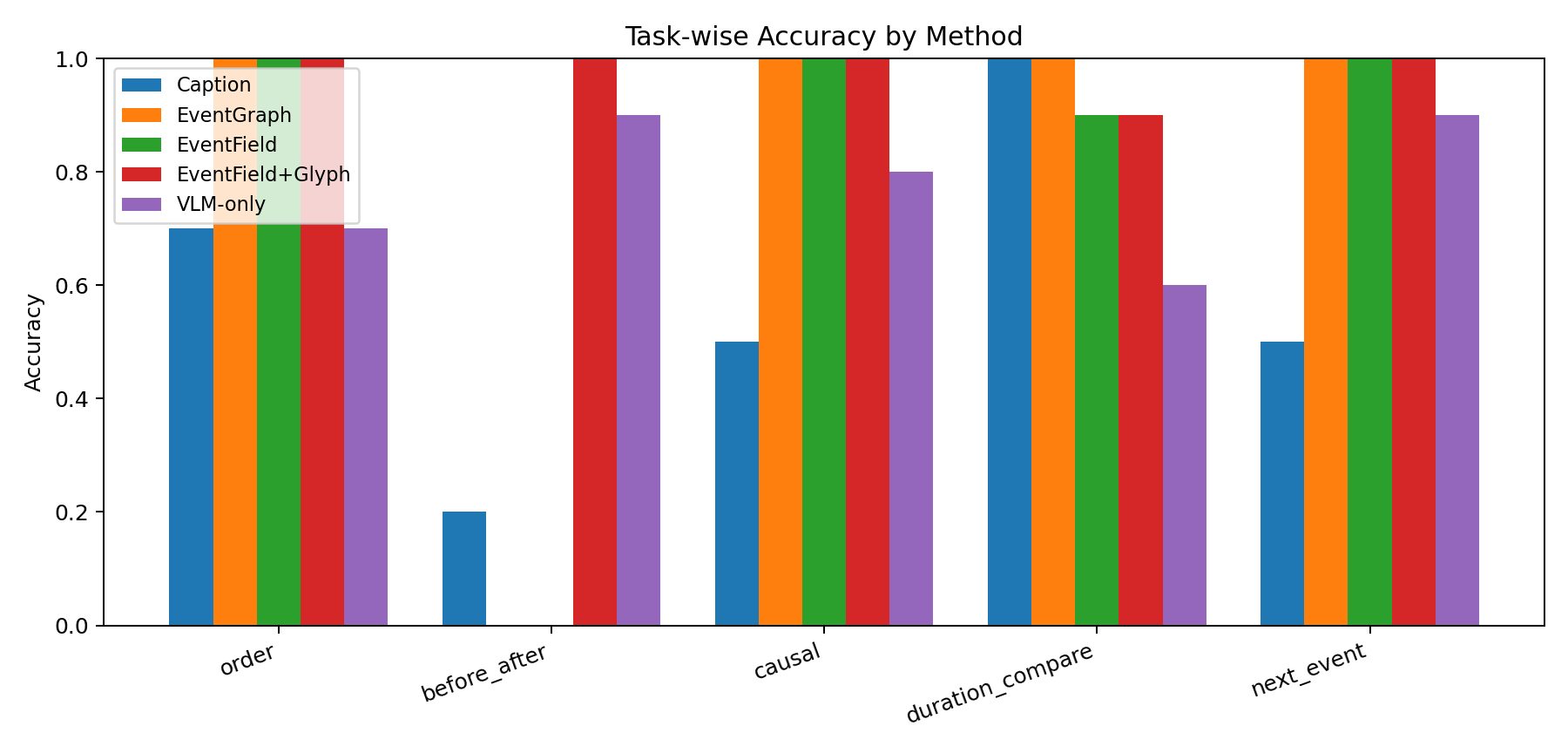}
  \caption{Task-wise accuracy by method.}
  \label{fig:taskwise}
\end{figure}
Task-wise behavior suggests that structured representations mainly help order, causal, and next-event reasoning. The structured glyph variant shows the clearest gains on the before/after task in this subset.

\subsection{Annotation-source variations}
\begin{table*}[t]
\centering
\caption{Annotation-source variants on EPIC10 (10 videos, 50 tasks).}
\label{tab:auto_variants}
\resizebox{\textwidth}{!}{%
\begin{tabular}{l l c c c c c c}
\toprule
Source & Best method & Best acc. & Caption acc. & $\Delta$ vs caption & p-value & Avg events/video & Avg canonical actions/video \\
\midrule
manual & eventfield\_glyph & 0.980 & 0.580 & 0.400 & 0.0000 & 35.2 & 24.7 \\
heuristic & eventfield\_glyph & 0.940 & 0.660 & 0.280 & 0.0026 & 113.6 & 2.0 \\
heuristic\_gemini & eventfield\_glyph & 0.920 & 0.720 & 0.200 & 0.0309 & 46.3 & 14.2 \\
\bottomrule
\end{tabular}
}
\end{table*}
Across manual, heuristic, and heuristic+Gemini sources, the best-performing method remains a structured EventField+Glyph variant. This is consistent with robust downstream reasoning as annotation density and quality change.
The table also shows that the absolute numbers shift with annotation quality, but the ordering remains stable, which is the main sign that the method is not specific to one annotation pipeline. In other words, the structured representation continues to support the comparison even when the upstream event inventory becomes noisier or denser.
We treat this as a robustness check rather than as proof that the method is insensitive to all annotation noise.

\subsection{Cross-video pair benchmark}
{\centering
\small
\captionof{table}{Cross-video pair benchmark (task set v2).}
\label{tab:pair_v2}
\resizebox{0.75\columnwidth}{!}{%
\begin{tabular}{lccc}
\toprule
Method & Overall Acc. & Macro Avg. & N \\
\midrule
caption & 0.694 & 0.694 & 36 \\
event\_graph & 0.917 & 0.917 & 36 \\
eventfield & 0.889 & 0.889 & 36 \\
eventfield\_glyph & 0.944 & 0.944 & 36 \\
\bottomrule
\end{tabular}
}
\par}
The pair benchmark shows the same ranking trend as the single-video tasks: caption is weakest, while structured methods are strongest and better aligned with temporal-difference reasoning.
This is consistent with the main results table and reinforces the same qualitative pattern across both single-video and pairwise settings. Even when the task changes from within-clip reasoning to cross-clip comparison, the structured representation remains a strong option.
Taken together, these results are best read as evidence that the representation is useful on this curated benchmark, not as a claim that it dominates all video reasoning settings.

%% file: tables/main_results.tex
\begin{table}[t]
\centering
\caption{Main method comparison on EPIC10 calibrated subset.}
\label{tab:main_results}
\resizebox{\columnwidth}{!}{%
\begin{tabular}{lccc}
\toprule
Method & Overall Acc. & Macro Avg. & N \\
\midrule
caption & 0.580 & 0.580 & 50 \\
event\_graph & 0.800 & 0.800 & 50 \\
eventfield & 0.780 & 0.780 & 50 \\
eventfield\_glyph & 0.980 & 0.980 & 50 \\
vlm\_only & 0.780 & 0.780 & 50 \\
\bottomrule
\end{tabular}
}
\end{table}

%% file: tables/pairwise_significance.tex
\begin{table}[t]
\centering
\caption{Pairwise significance (paired bootstrap CI and McNemar exact p-value).}
\label{tab:pairwise}
\resizebox{\columnwidth}{!}{%
\begin{tabular}{lccc}
\toprule
Pair & $\Delta$ Acc. (A-B) & 95\% CI & p-value \\
\midrule
caption vs event\_graph & -0.220 & [-0.360, -0.080] & 0.0074 \\
caption vs eventfield & -0.200 & [-0.340, -0.060] & 0.0213 \\
caption vs eventfield\_glyph & -0.400 & [-0.540, -0.240] & 0.0000 \\
caption vs vlm\_only & -0.200 & [-0.380, -0.020] & 0.0525 \\
event\_graph vs eventfield & 0.020 & [0.000, 0.060] & 1.0000 \\
event\_graph vs eventfield\_glyph & -0.180 & [-0.300, -0.060] & 0.0117 \\
event\_graph vs vlm\_only & 0.020 & [-0.160, 0.200] & 1.0000 \\
eventfield vs eventfield\_glyph & -0.200 & [-0.320, -0.100] & 0.0020 \\
eventfield vs vlm\_only & 0.000 & [-0.180, 0.180] & 1.0000 \\
eventfield\_glyph vs vlm\_only & 0.200 & [0.080, 0.320] & 0.0063 \\
\bottomrule
\end{tabular}
}
\end{table}

%% file: sec/5_conclusion.tex
\section{Conclusion}
\label{sec:conclusion}

We presented an interpretable temporal reasoning stack that combines EventGraph, EventField, and Glyph. On the current EPIC subset, structured methods outperform the caption baseline, and the strongest EventField+Glyph variant achieves the highest observed score. Source-variation studies suggest that the representation-centric approach remains useful under changing annotation quality, while the glyph view provides a practical diagnostic layer for human inspection.

The current evidence is limited to 10 videos and 50 questions, and the pipeline still depends on event annotations and largely rule-based reasoning modules. Future work should expand the evaluation scale, strengthen learned proposal and prediction modules, broaden cross-video comparison with higher statistical power, and add a formal human-subject study of interpretability.
Even with those limits, the current results suggest that the representation may be useful as a debugging and reasoning aid, which makes it a practical base for a larger learned system.

%% file: sec/6_limitations.tex
\section{Limitations}
This study is limited to a calibrated EPIC-KITCHENS subset of 10 videos and 50 questions, so the reported effect sizes should be read as evidence on a focused benchmark rather than a general claim across all video reasoning settings.

The pipeline also still depends on event annotations and mostly rule-based reasoning modules, and we do not yet include a formal human-subject study of interpretability. As a result, the glyph view should be understood as a promising diagnostic interface supported by qualitative examples and quantitative performance gains, not as a fully user-validated explanation system.

%% file: sec/7_reproducibility_summary.tex
\section{Reproducibility Workflow Summary}
All reported numbers were generated from version-controlled artifacts and exported into publication-ready tables and figures.

%% file: sec/6_appendix_eventglyphs.tex
\section{EventGlyph Gallery}
This supplementary material includes all EPIC10 EventGlyph outputs used in the study.

\newcommand{\GlyphRow}[3]{%
  \begin{subfigure}[t]{0.48\textwidth}
  \centering
  \includegraphics[width=\linewidth,height=0.20\textheight,keepaspectratio]{figures/appendix_glyphs/#2}
  \end{subfigure}
  \hfill
  \begin{subfigure}[t]{0.48\textwidth}
  \centering
  \includegraphics[width=\linewidth,height=0.20\textheight,keepaspectratio]{figures/appendix_glyphs/#3}
  \end{subfigure}
  \par\smallskip
  \noindent\textit{#1}\par\smallskip
}

\begin{figure*}[t]
\centering
\GlyphRow{P01\_13}{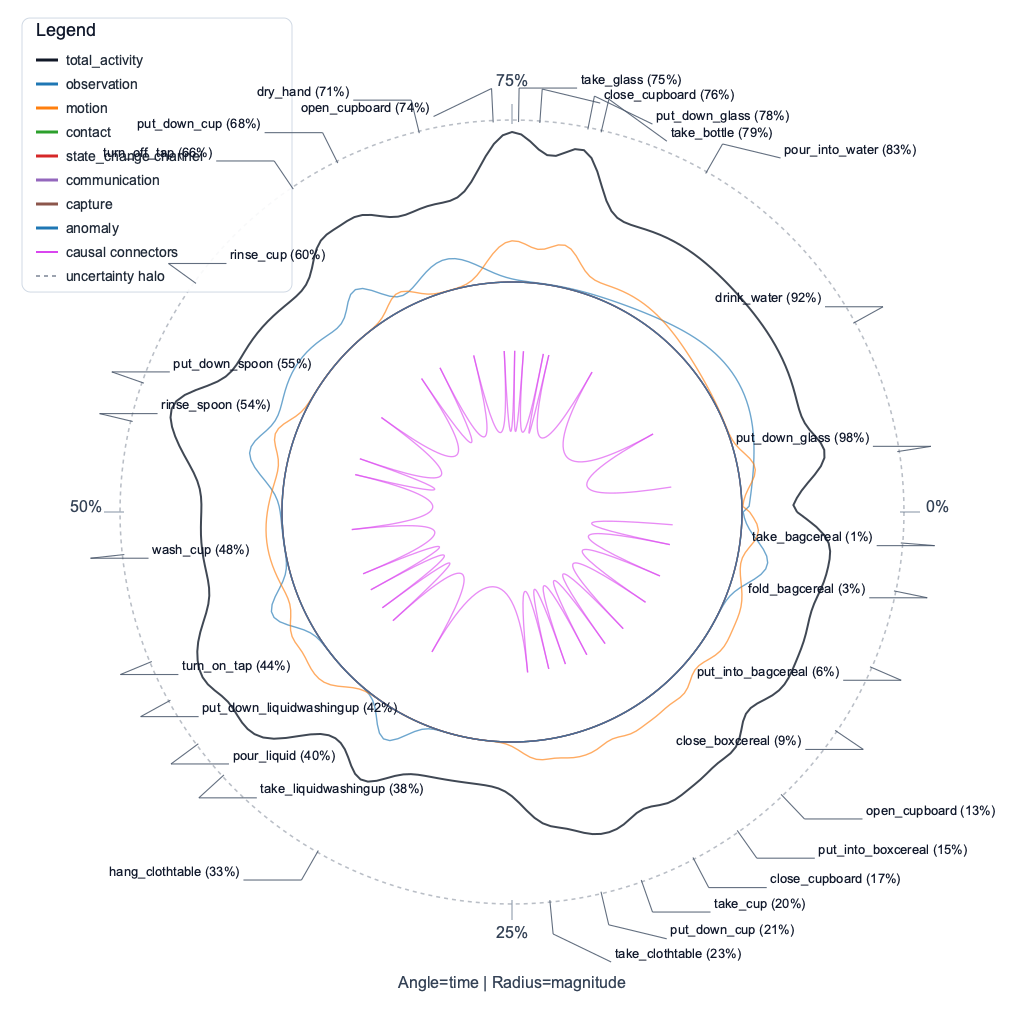}{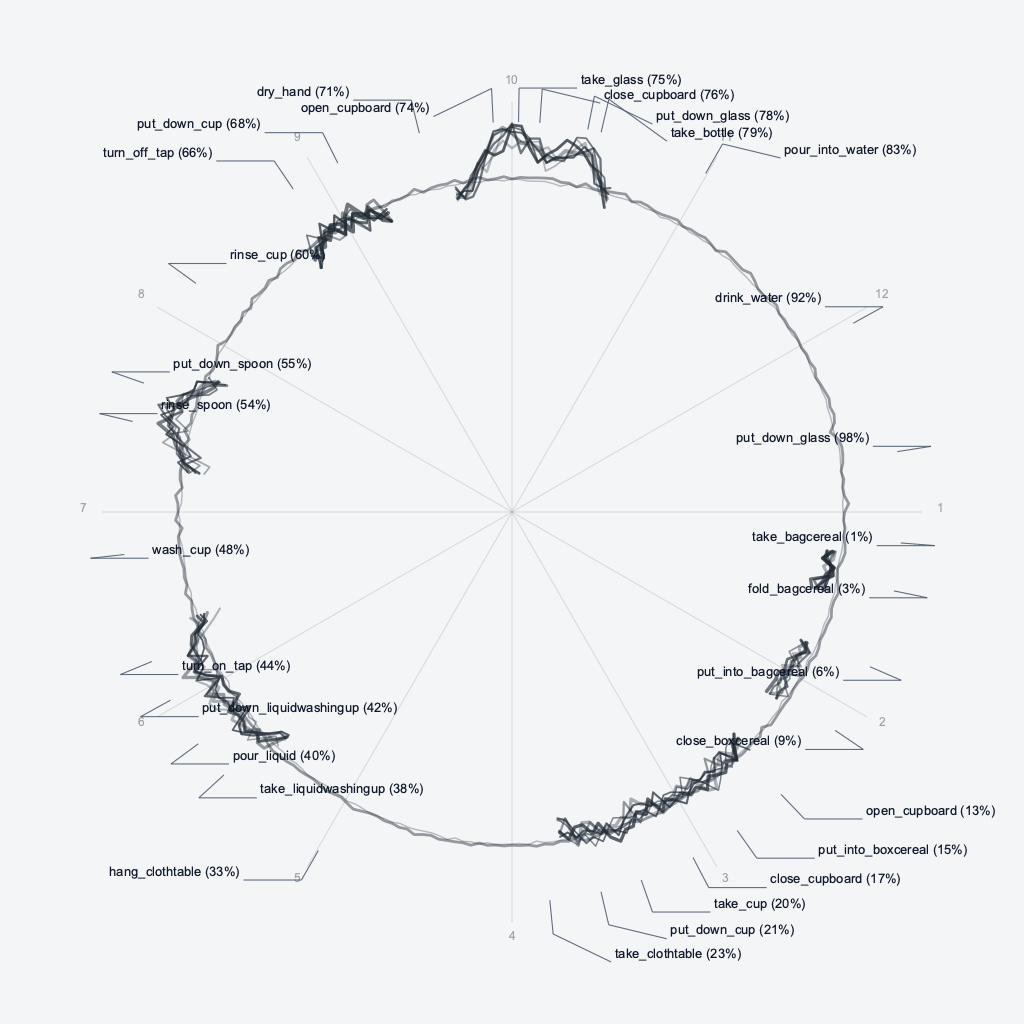}
\GlyphRow{P02\_15}{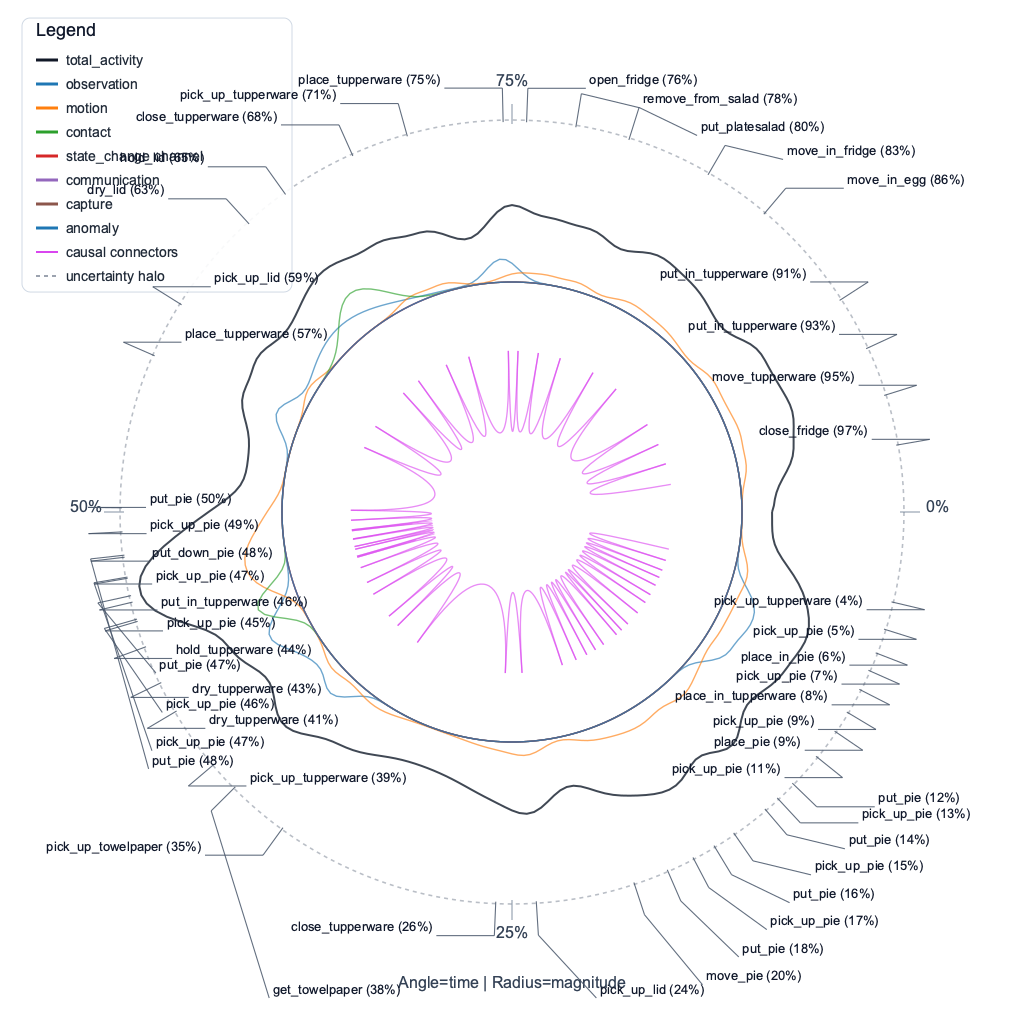}{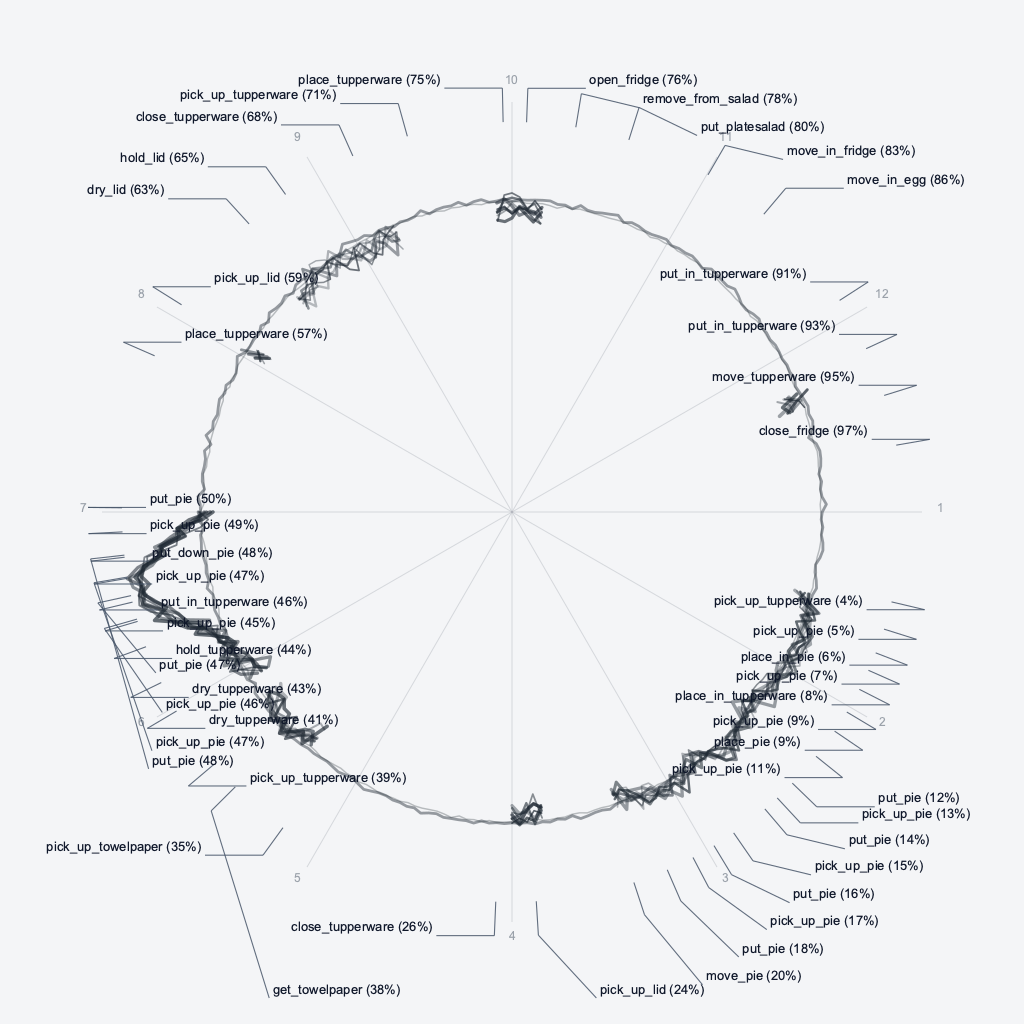}
\GlyphRow{P06\_12}{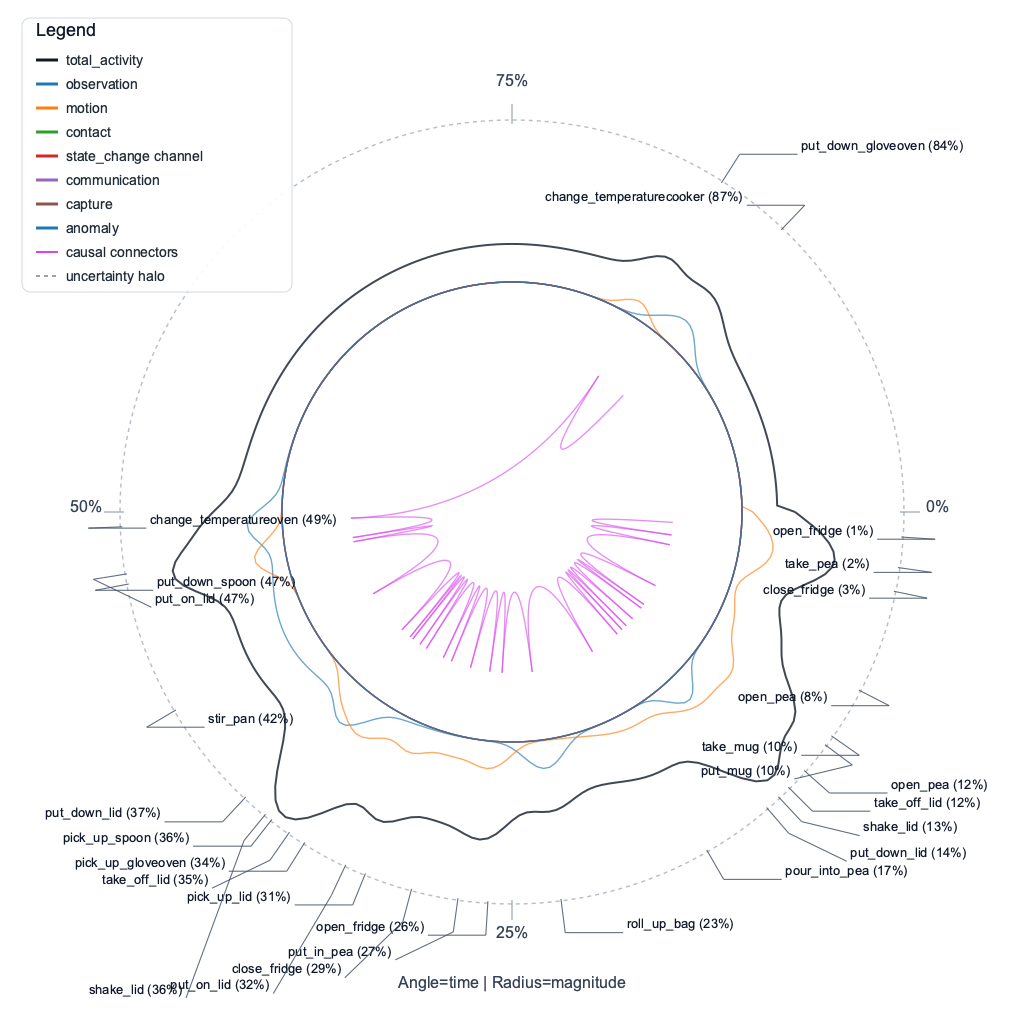}{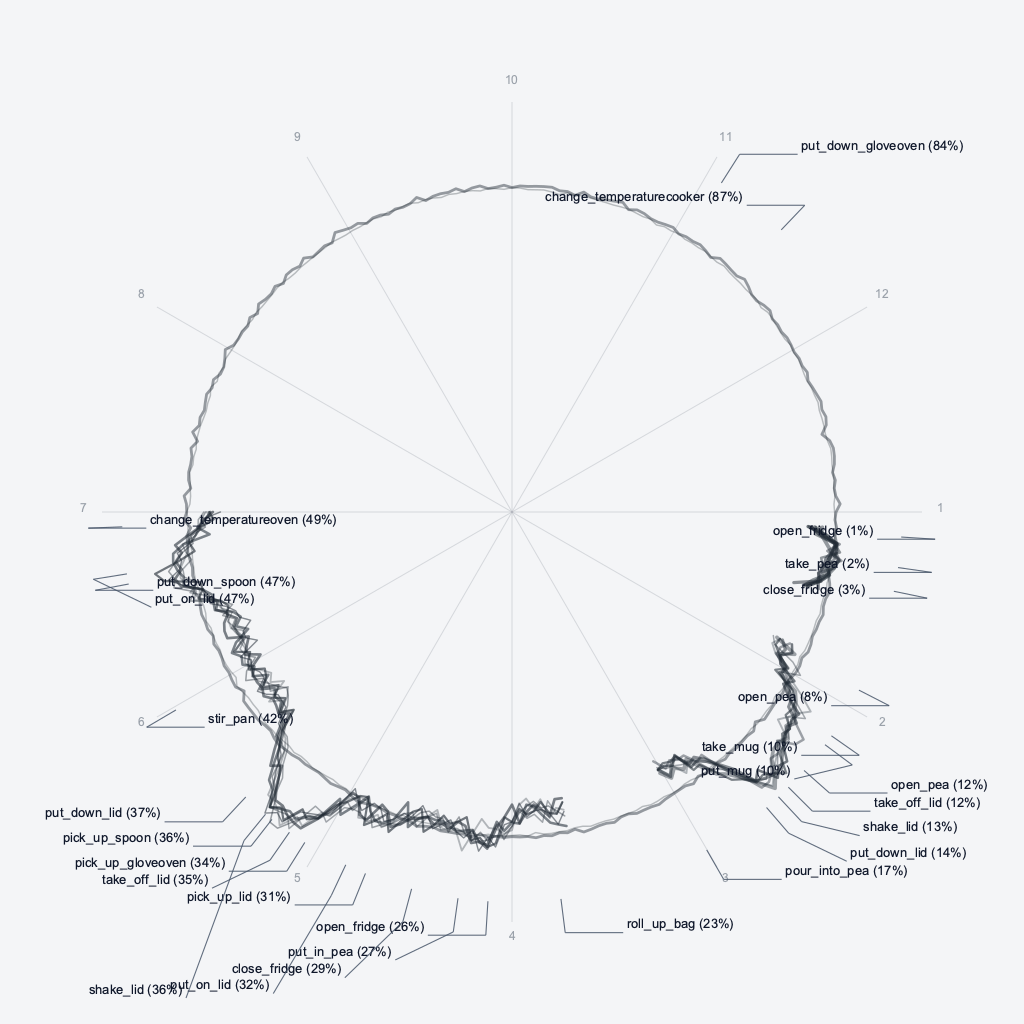}
\caption{EventGlyph gallery page 1: P01\_13, P02\_15, and P06\_12.}
\end{figure*}

\begin{figure*}[t]
\centering
\GlyphRow{P06\_14}{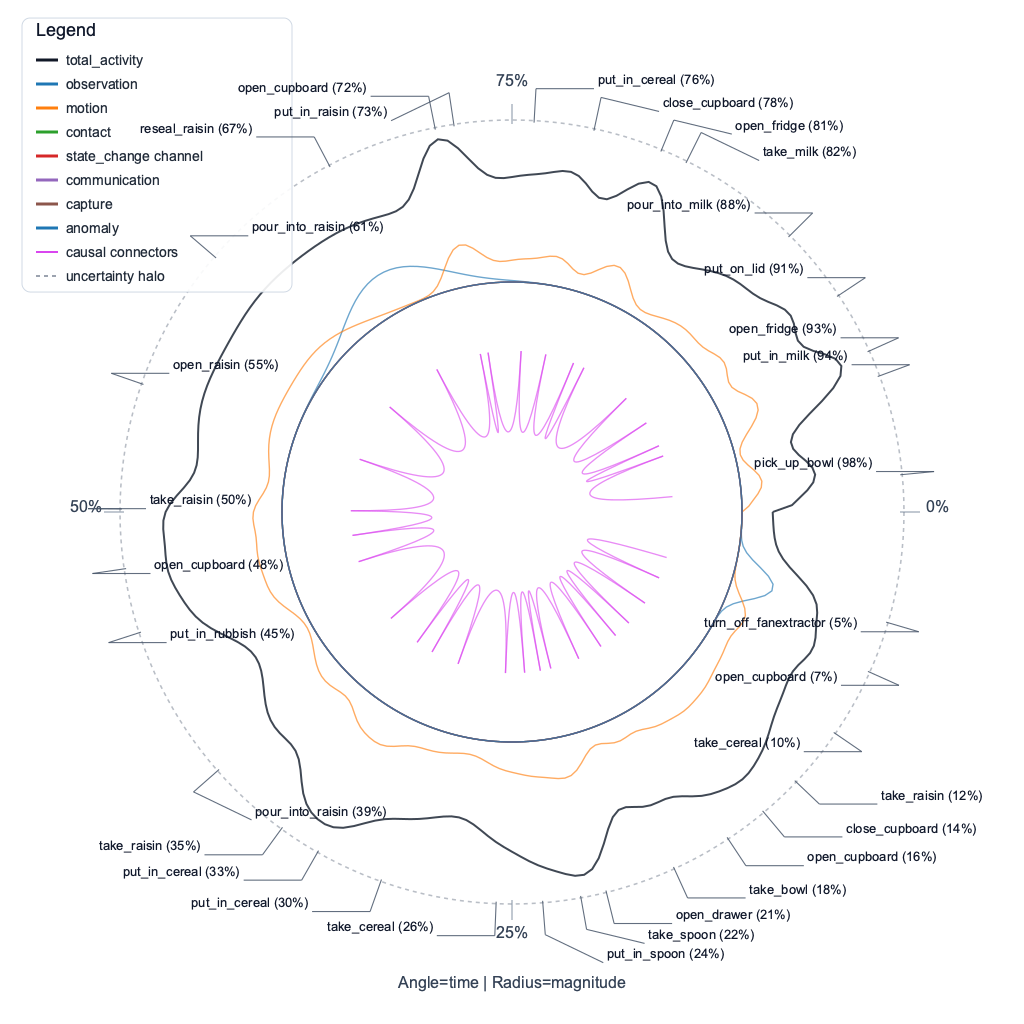}{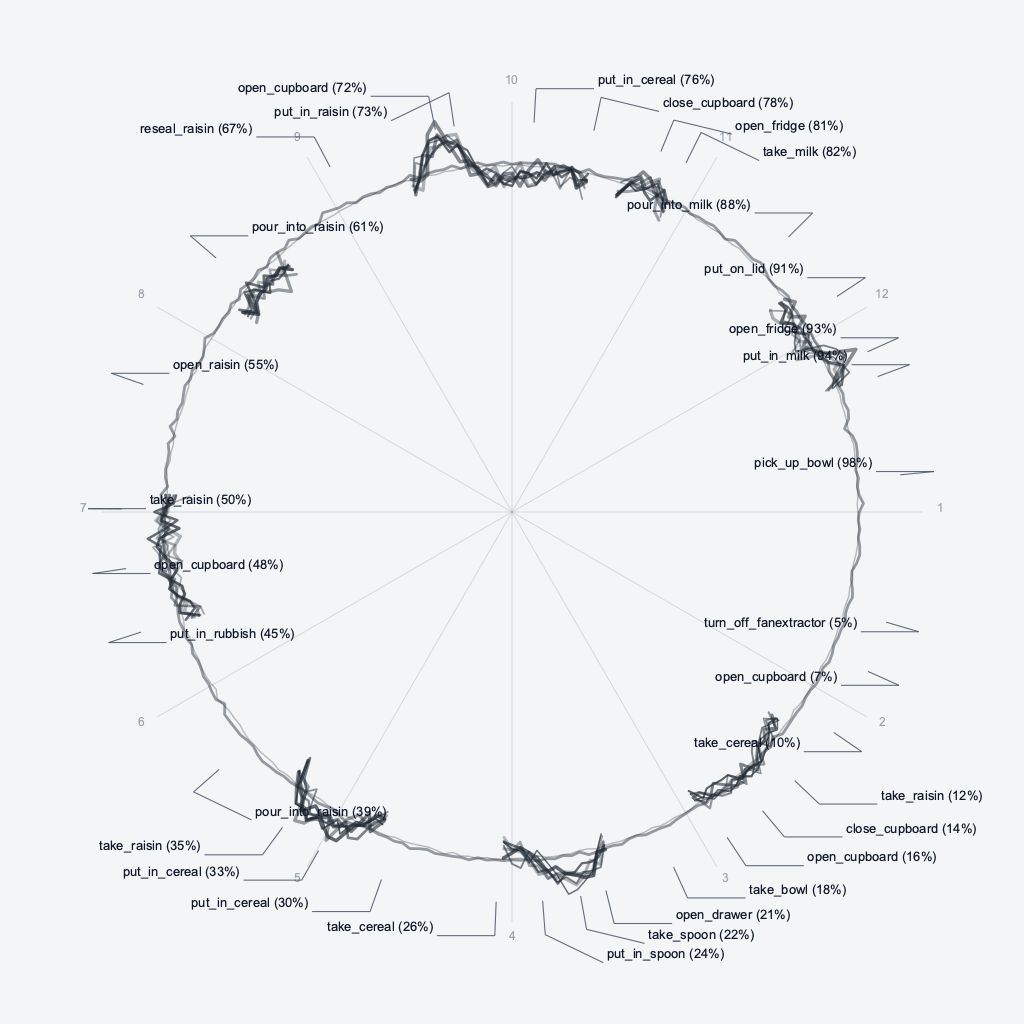}
\GlyphRow{P07\_12}{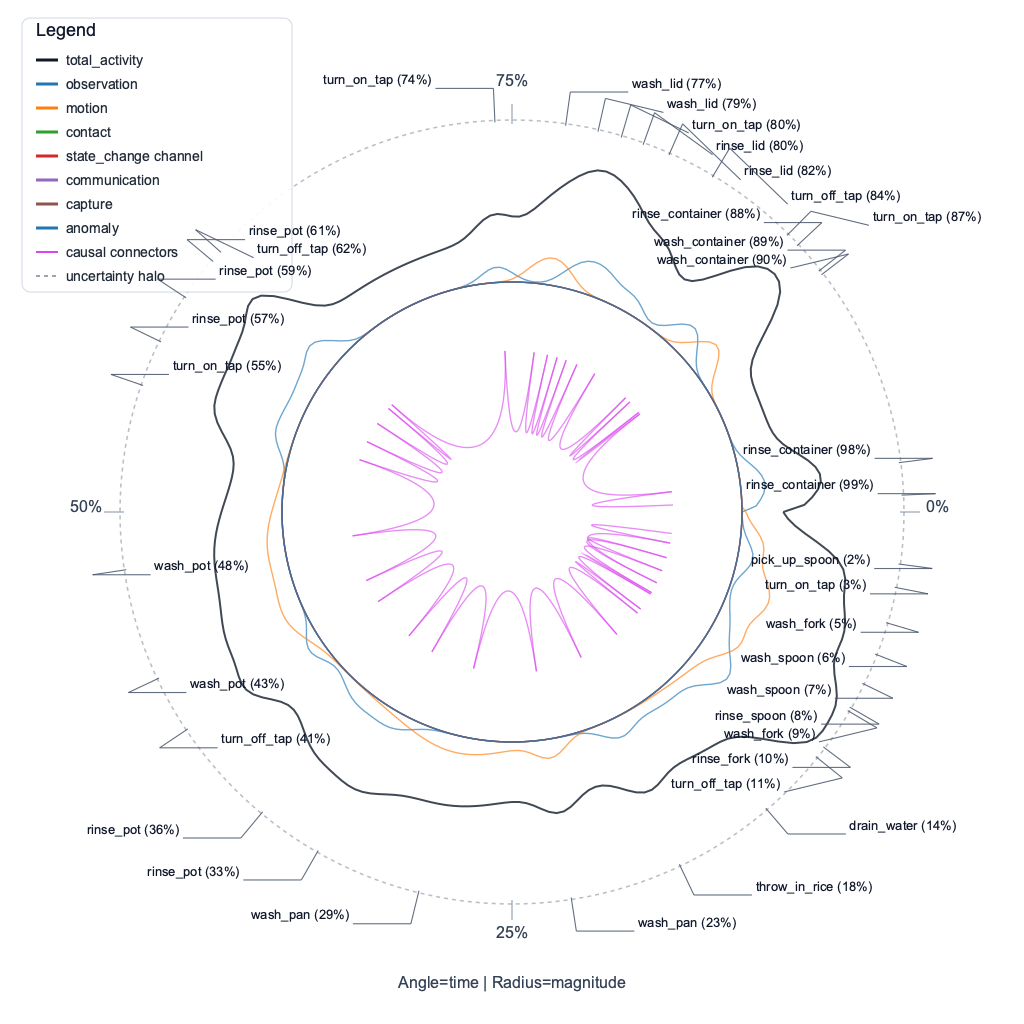}{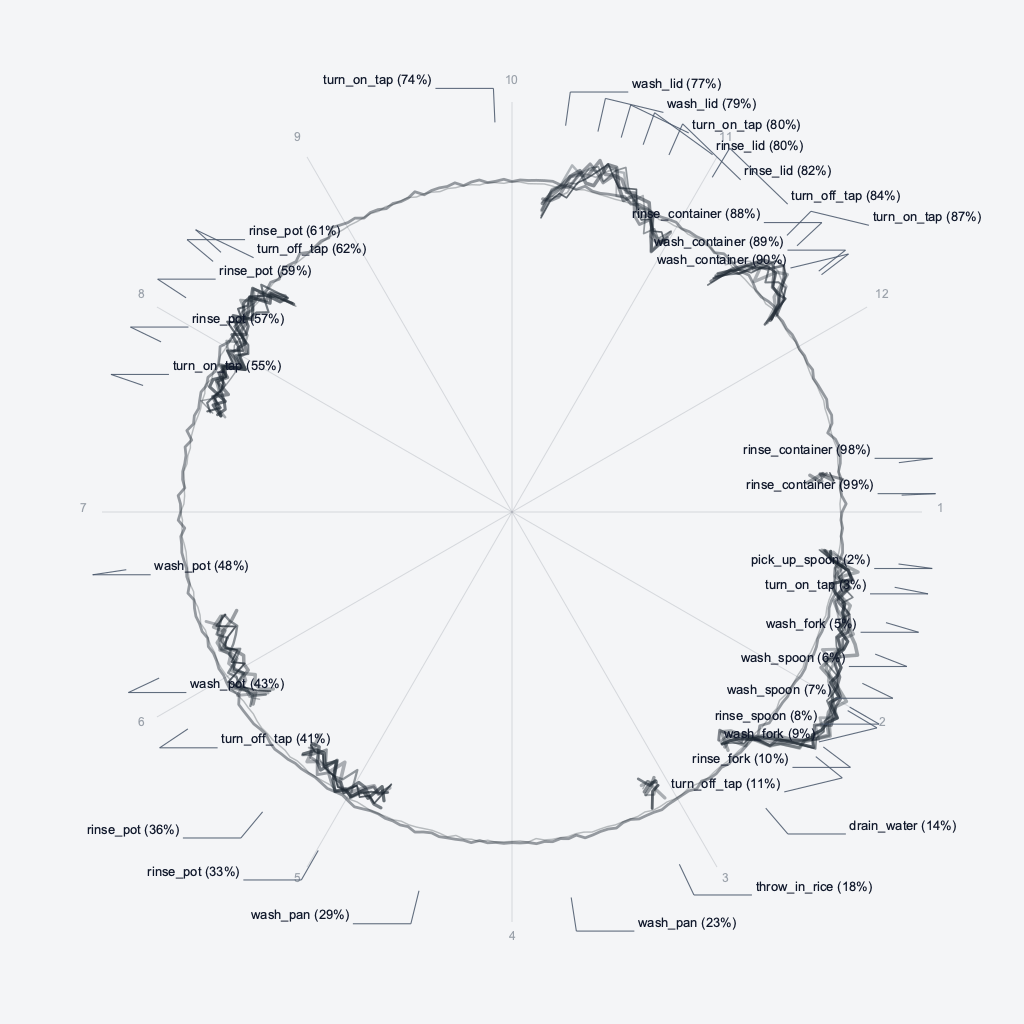}
\GlyphRow{P07\_16}{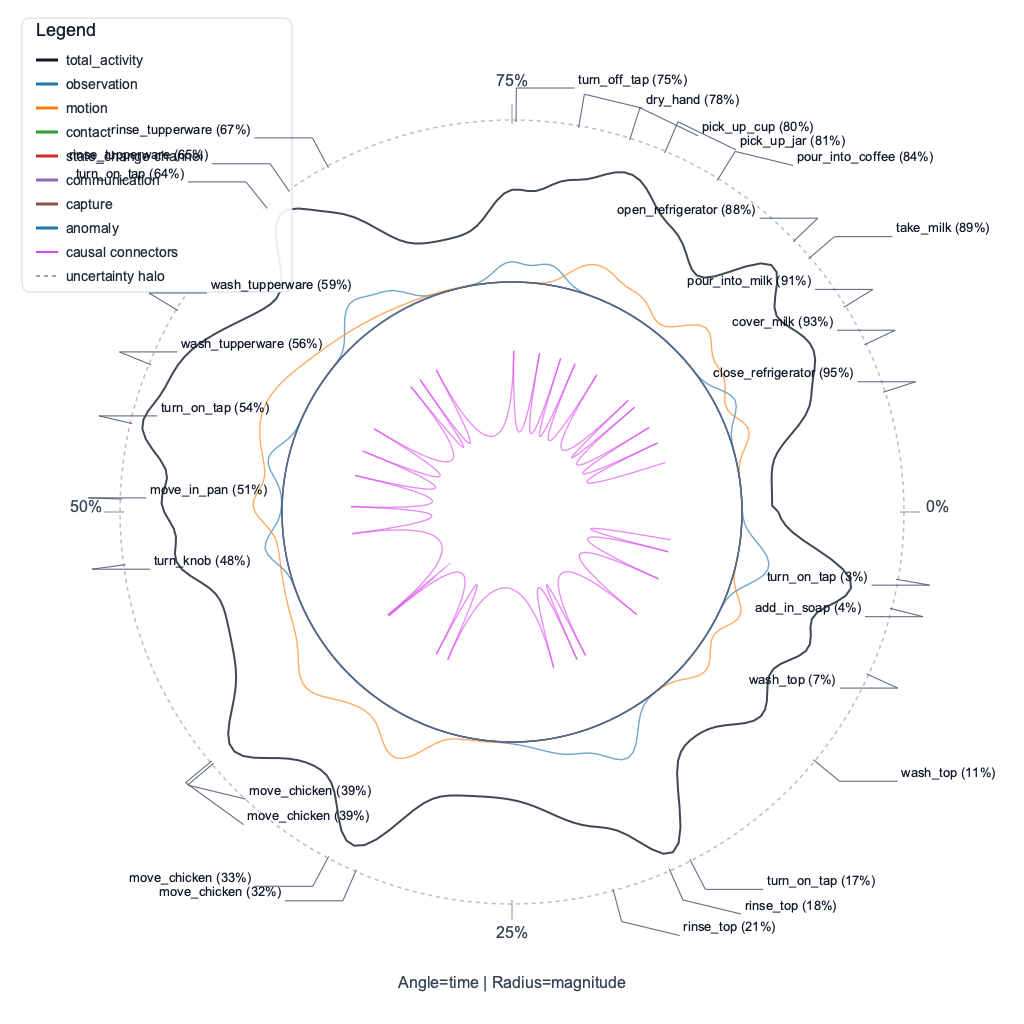}{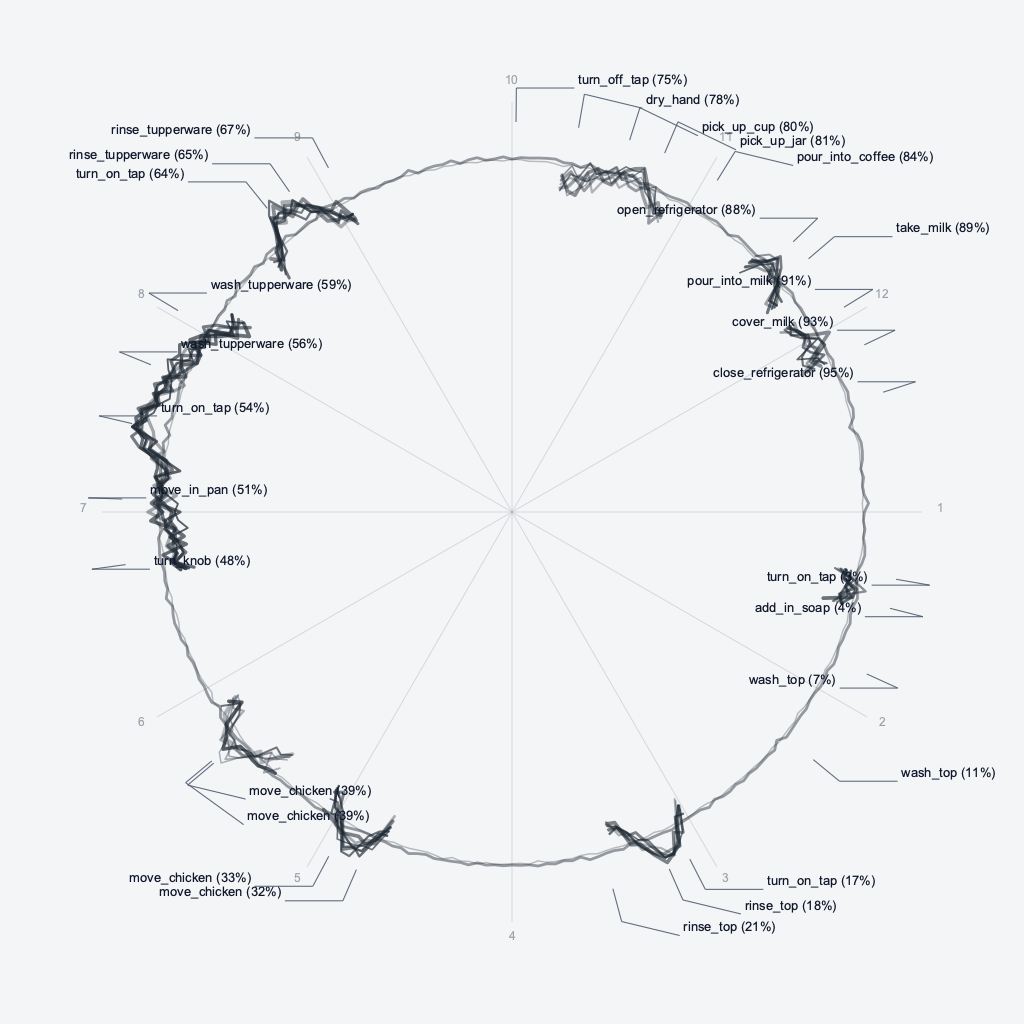}
\caption{EventGlyph gallery page 2: P06\_14, P07\_12, and P07\_16.}
\end{figure*}

\begin{figure*}[t]
\centering
\GlyphRow{P08\_14}{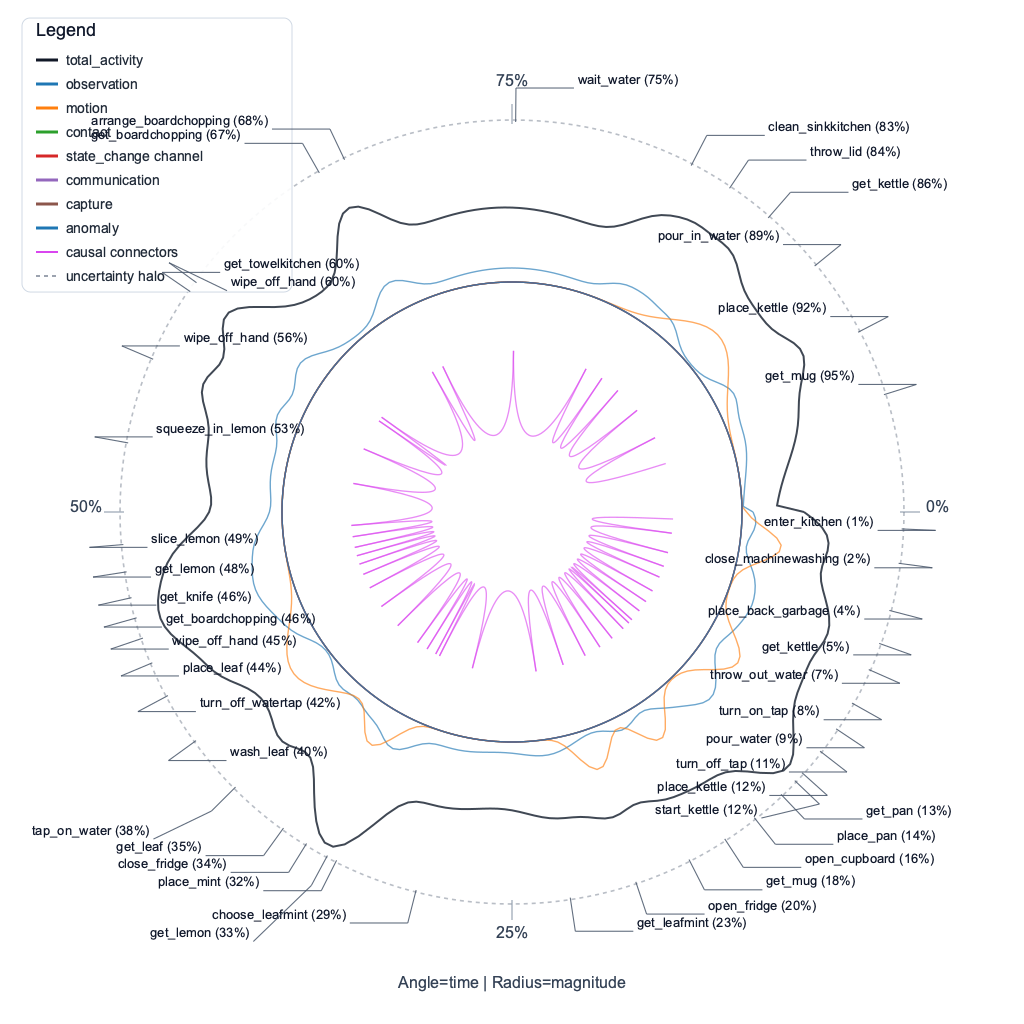}{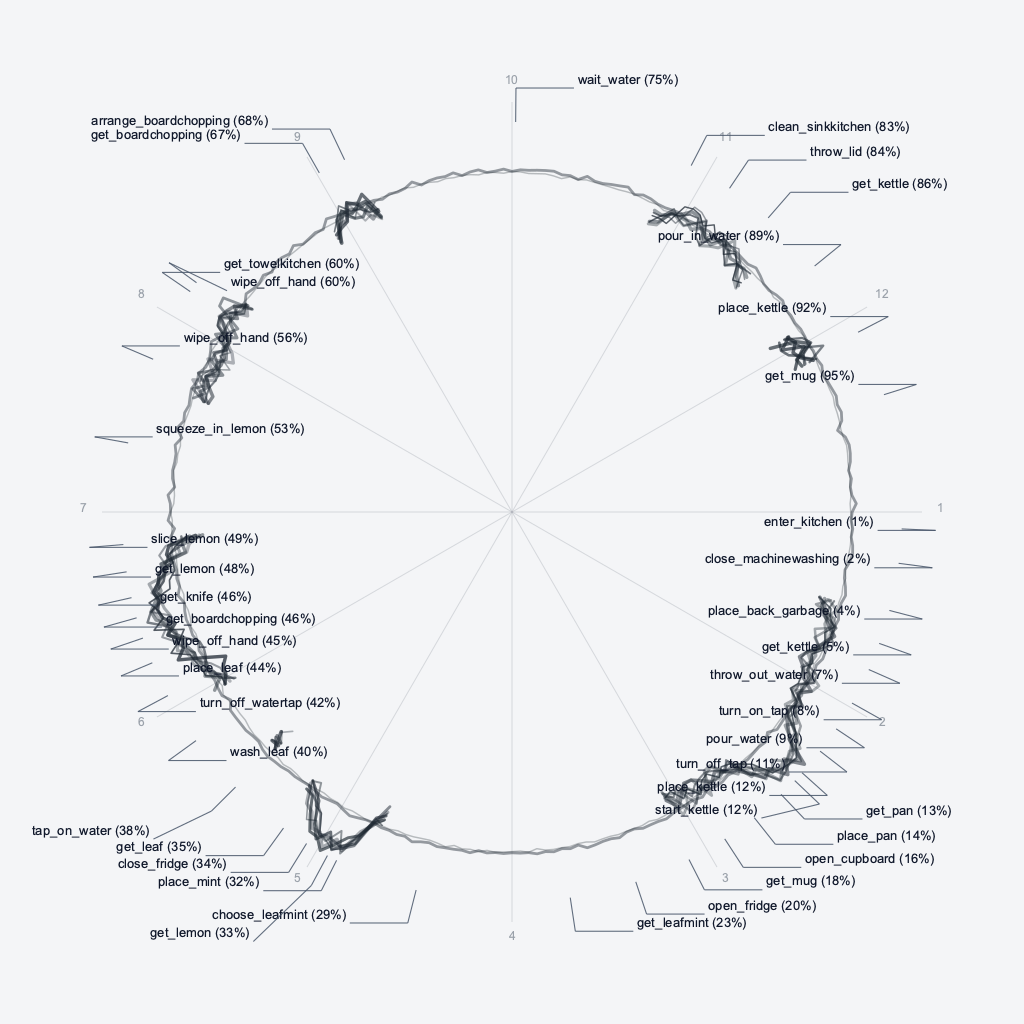}
\GlyphRow{P09\_08}{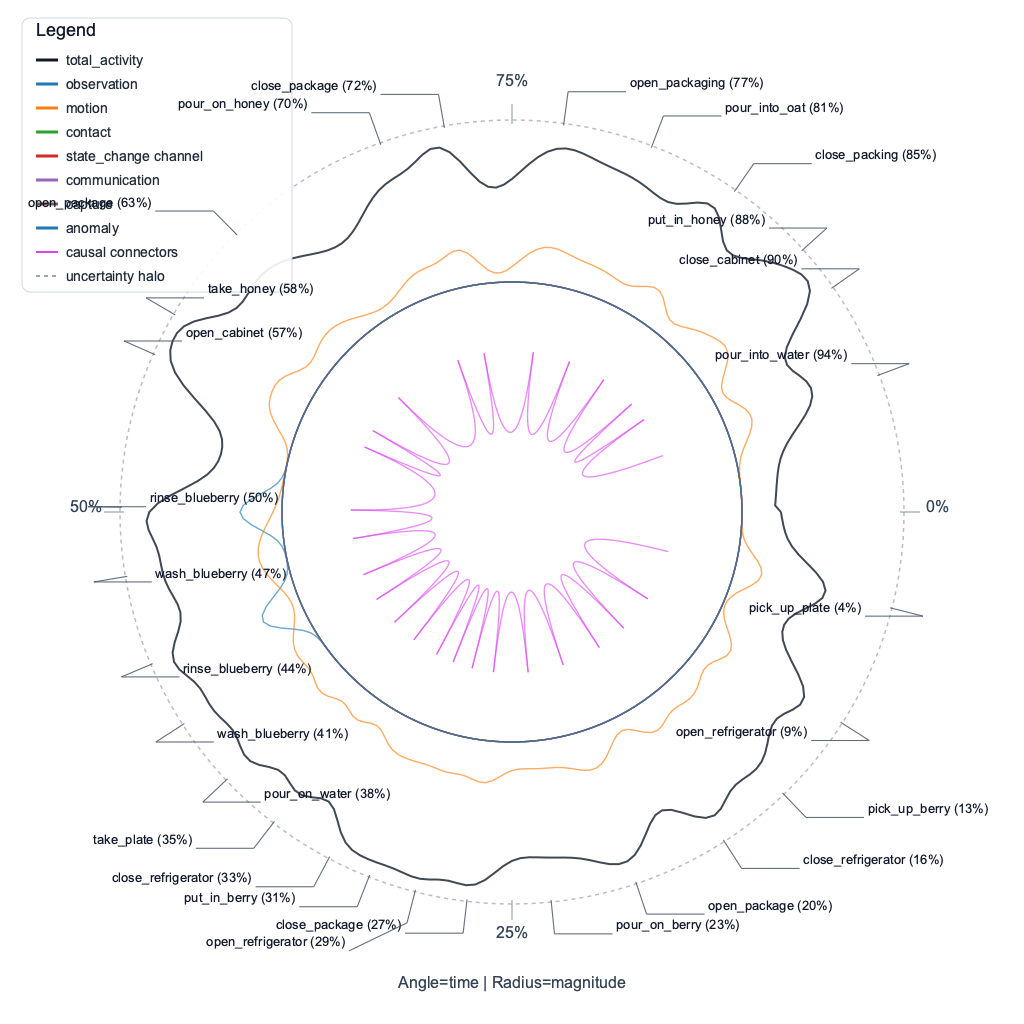}{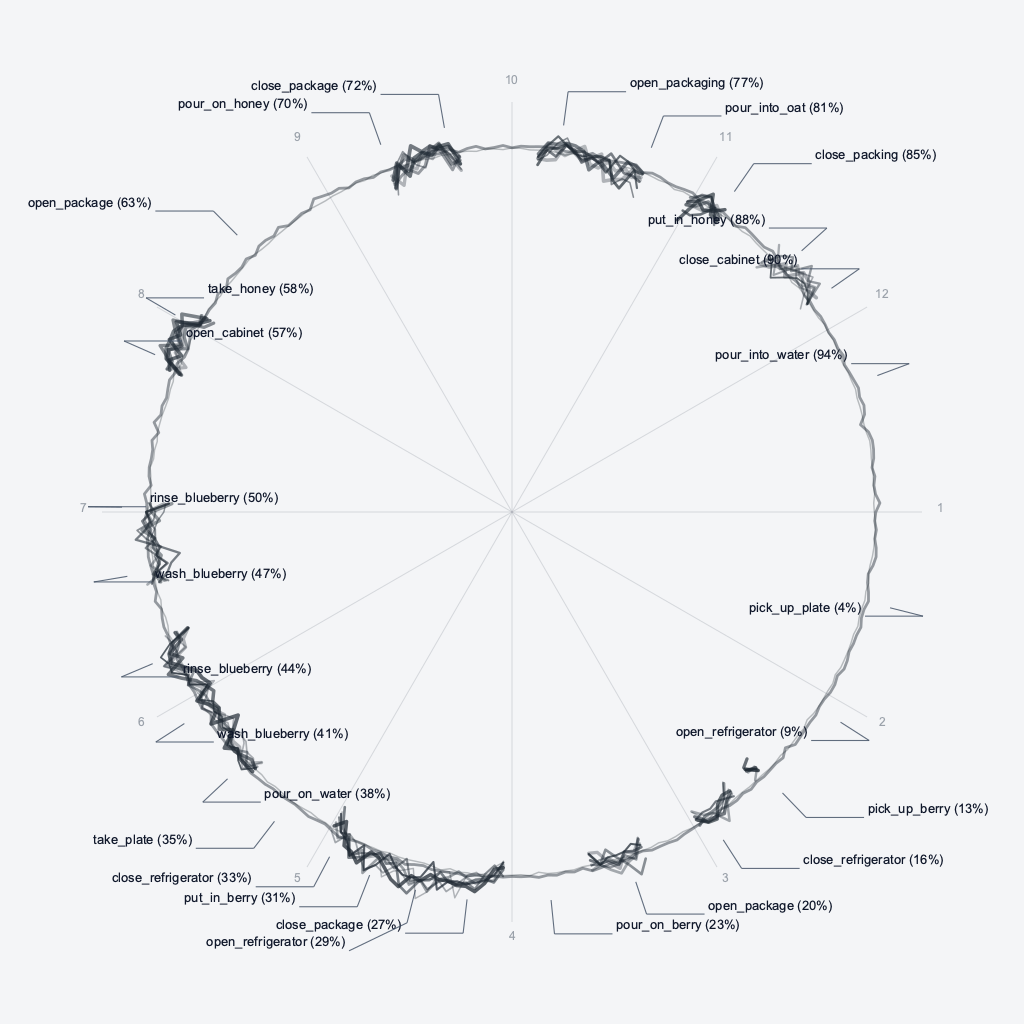}
\caption{EventGlyph gallery page 3: P08\_14 and P09\_08.}
\end{figure*}

\begin{figure*}[t]
\centering
\GlyphRow{P11\_19}{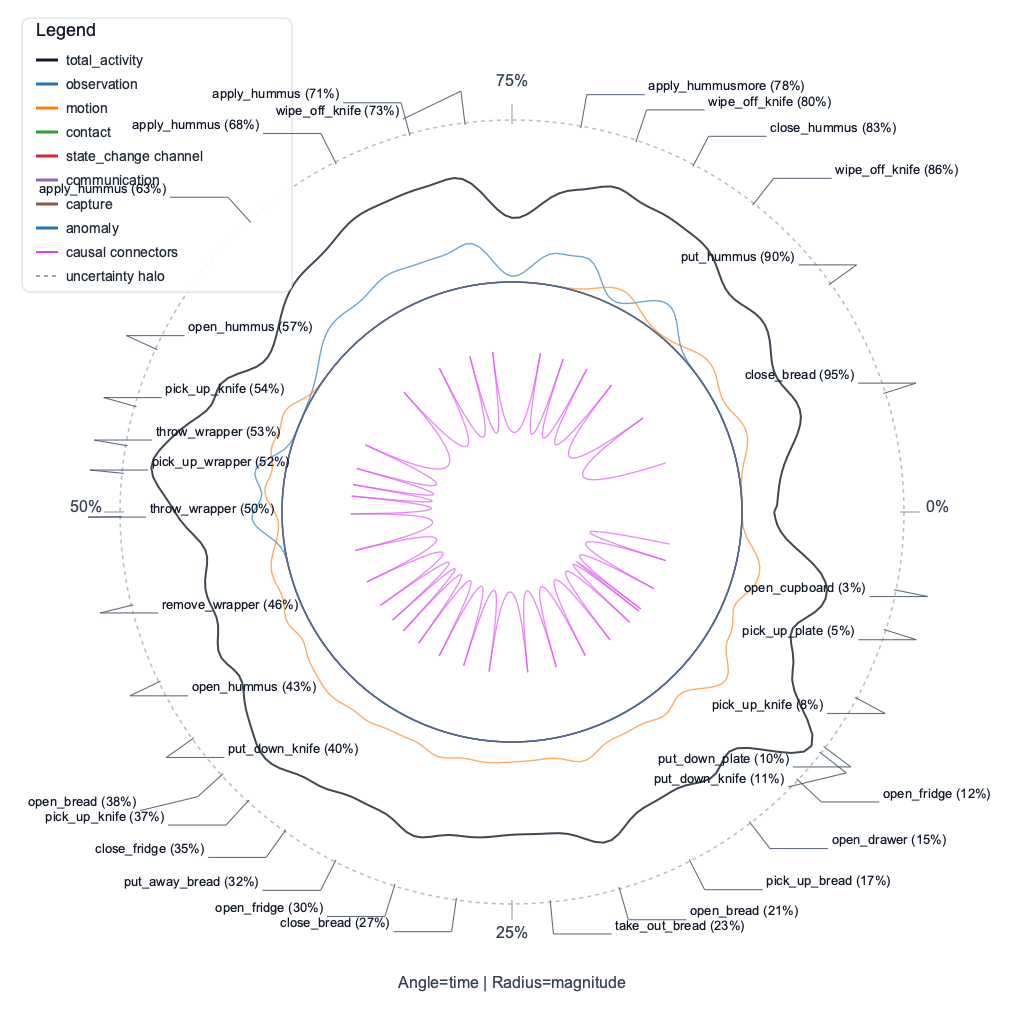}{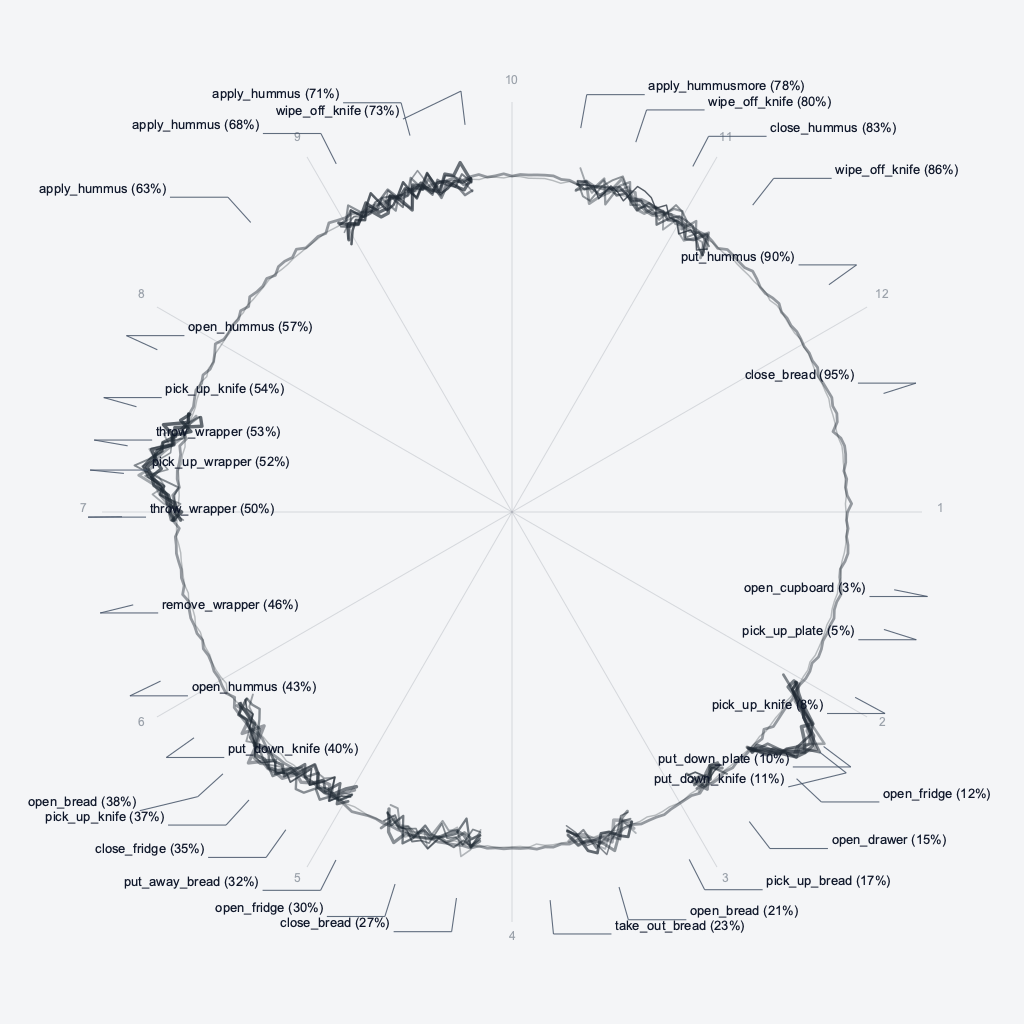}
\GlyphRow{P18\_01}{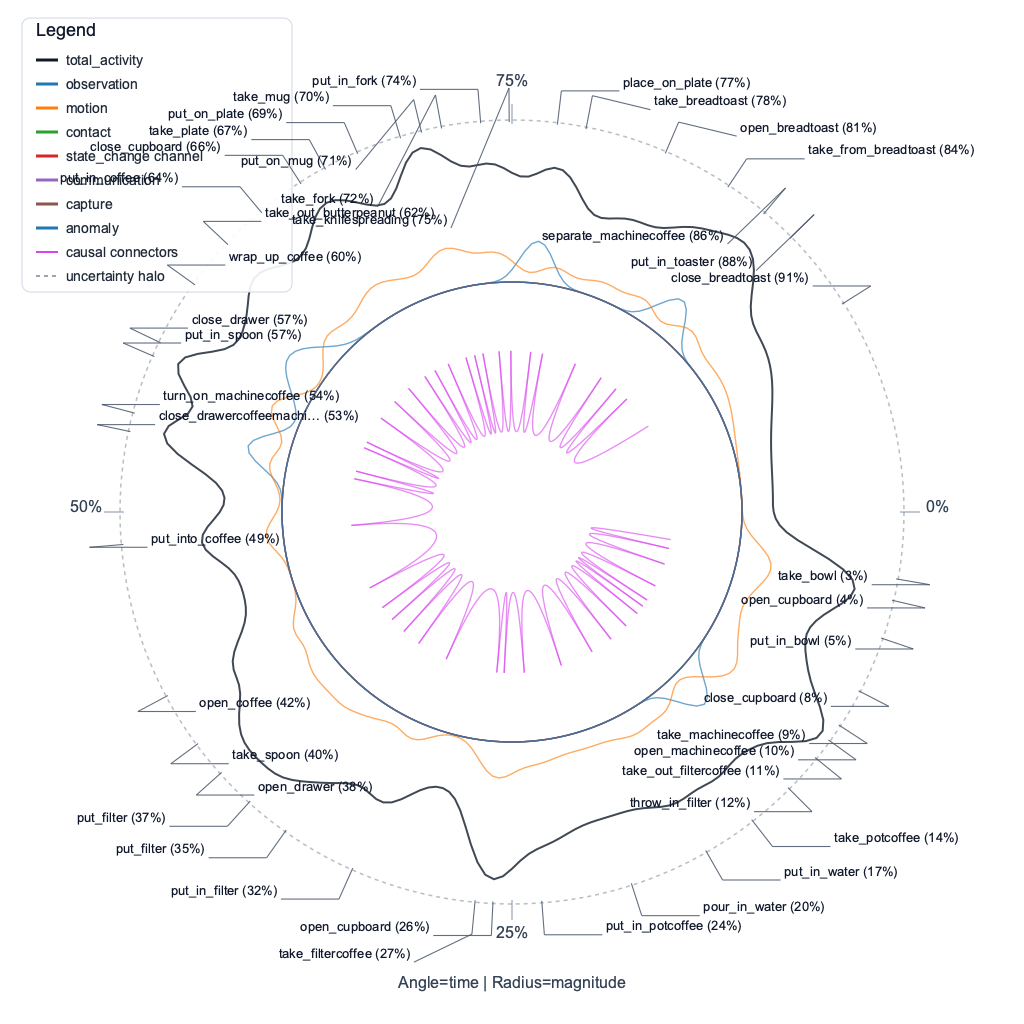}{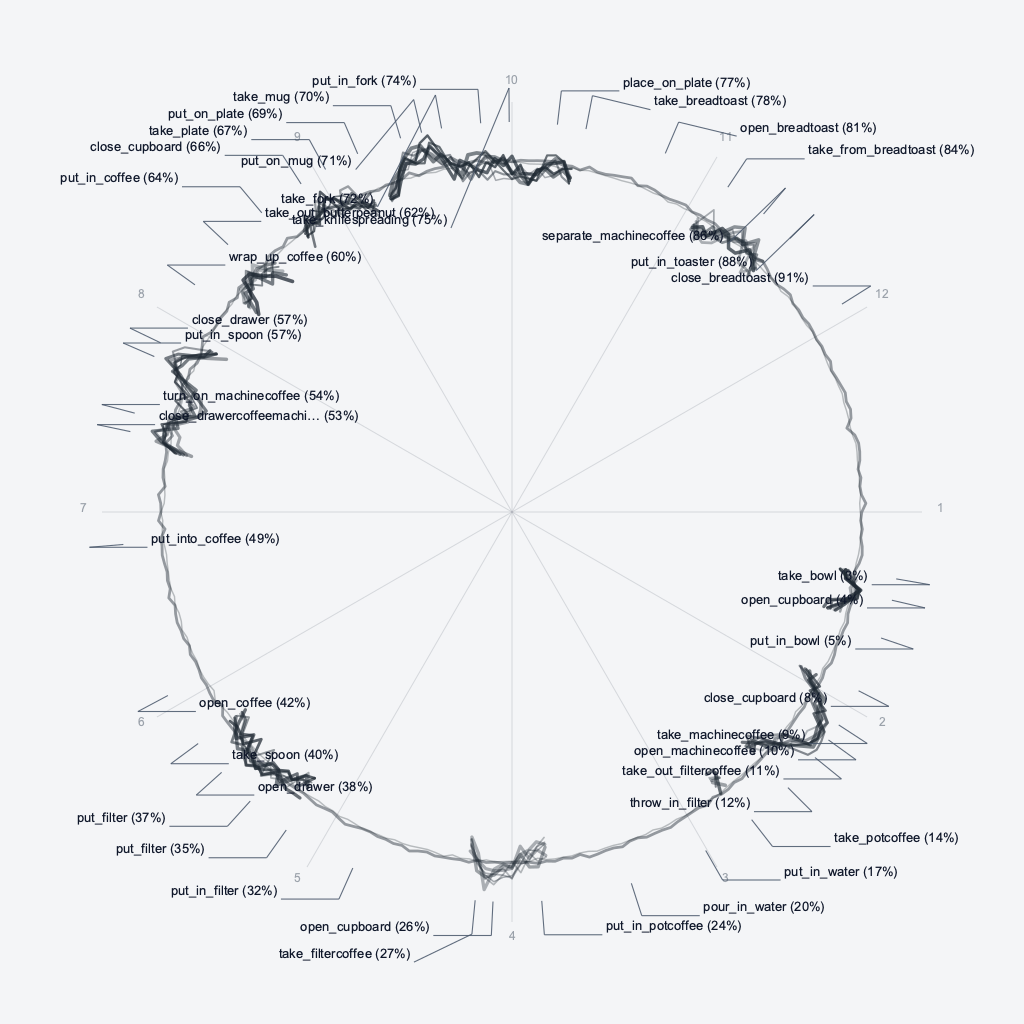}
\caption{EventGlyph gallery page 4: P11\_19 and P18\_01.}
\end{figure*}

%% file: main.bib
@inproceedings{damen2020epic,
  title={Rescaling Egocentric Vision: Collection, Pipeline and Challenges for EPIC-KITCHENS-100},
  author={Damen, Dima and Doughty, Hazel and Farinella, Giovanni Maria and Furnari, Antonino and Kazakos, Evangelos and Moltisanti, Davide and Munro, Jonathan and Perrett, Toby and Price, Will and Wray, Michael},
  booktitle={International Journal of Computer Vision},
  year={2022}
}

@inproceedings{lei2018tvqa,
  title={TVQA: Localized, Compositional Video Question Answering},
  author={Lei, Jie and Yu, Licheng and Bansal, Mohit and Berg, Tamara},
  booktitle={EMNLP},
  year={2018}
}

@inproceedings{grunde-mclaughlin2021agqa,
  title={AGQA: A Benchmark for Compositional Spatio-Temporal Reasoning},
  author={Grunde-McLaughlin, Madeleine and Krishna, Ranjay and Agrawala, Maneesh},
  booktitle={Proceedings of the IEEE/CVF Conference on Computer Vision and Pattern Recognition (CVPR)},
  year={2021},
  pages={11287-11297}
}

@inproceedings{yu2023anetqa,
  title={ANetQA: A Large-Scale Benchmark for Fine-Grained Compositional Reasoning Over Untrimmed Videos},
  author={Yu, Zhou and Zheng, Lixiang and Zhao, Zhou and Wu, Fei and Fan, Jianping and Ren, Kui and Yu, Jun},
  booktitle={Proceedings of the IEEE/CVF Conference on Computer Vision and Pattern Recognition (CVPR)},
  year={2023},
  pages={23191-23200}
}

@inproceedings{jang2017tgifqa,
  title={TGIF-QA: Toward Spatio-Temporal Reasoning in Visual Question Answering},
  author={Jang, Yohan and Song, Yunseok and Yu, Youngjae and Kim, Youngbin and Kim, Gunhee},
  booktitle={CVPR},
  year={2017}
}

@inproceedings{wang2018videos,
  title={Videos as Space-Time Region Graphs},
  author={Wang, Xiaolong and Gupta, Abhinav},
  booktitle={ECCV},
  year={2018}
}

@article{lipton2018mythos,
  title={The Mythos of Model Interpretability},
  author={Lipton, Zachary C.},
  journal={Queue},
  year={2018}
}

@article{rudin2019stop,
  title={Stop Explaining Black Box Machine Learning Models for High Stakes Decisions and Use Interpretable Models Instead},
  author={Rudin, Cynthia},
  journal={Nature Machine Intelligence},
  year={2019}
}

@inproceedings{min2024morevqa,
  title={MoReVQA: Exploring Modular Reasoning Models for Video Question Answering},
  author={Min, Juhong and Buch, Shyamal and Nagrani, Arsha and Cho, Minsu and Schmid, Cordelia},
  booktitle={Proceedings of the IEEE/CVF Conference on Computer Vision and Pattern Recognition (CVPR)},
  year={2024},
  pages={13235-13245}
}

@inproceedings{xiao2024grounded,
  title={Can I Trust Your Answer? Visually Grounded Video Question Answering},
  author={Xiao, Junbin and Yao, Angela and Li, Yicong and Chua, Tat-Seng},
  booktitle={Proceedings of the IEEE/CVF Conference on Computer Vision and Pattern Recognition (CVPR)},
  year={2024},
  pages={13204-13214}
}

@inproceedings{di2024groundvqa,
  title={Grounded Question-Answering in Long Egocentric Videos},
  author={Di, Shangzhe and Xie, Weidi},
  booktitle={Proceedings of the IEEE/CVF Conference on Computer Vision and Pattern Recognition (CVPR)},
  year={2024},
  pages={12934-12943}
}

@inproceedings{li2024mvbench,
  title={MVBench: A Comprehensive Multi-modal Video Understanding Benchmark},
  author={Li, Kunchang and Wang, Yali and He, Yinan and Li, Yizhuo and Wang, Yi and Liu, Yi and Wang, Zun and Xu, Jilan and Chen, Guo and Luo, Ping and Wang, Limin and Qiao, Yu},
  booktitle={Proceedings of the IEEE/CVF Conference on Computer Vision and Pattern Recognition (CVPR)},
  year={2024},
  pages={22195-22206}
}

@book{vanderaalst2016process,
  title={Process Mining: Data Science in Action},
  author={van der Aalst, Wil},
  publisher={Springer},
  year={2016}
}
